\documentclass[letterpaper, 10 pt, conference]{ieeeconf}  
\IEEEoverridecommandlockouts                              
\usepackage{graphics}    
\usepackage{times}       
\usepackage{amsmath}     
\usepackage{amssymb}     
\usepackage{graphicx}
\usepackage{algorithm}
\usepackage[noend]{algpseudocode}
\usepackage{xspace}
\usepackage{multirow}
\usepackage{cite}
\usepackage{xcolor}
\usepackage[font=small]{caption}

\def\secref#1{Sec.~\ref{#1}}
\def\figref#1{Fig.~\ref{#1}}

\def\eqref#1{Eq.~(\ref{#1})}

\newcommand{\ie}{i.e.,\xspace}

\newcommand{\etalcite}[1]{\emph{et al.}~\cite{#1}}
\newcommand{\ours}{\mbox{RHINO-AR}\xspace}
\newcommand{\vr}{\mbox{RHINO-VR}\xspace}

\title{\LARGE \bf RHINO-AR: An Augmented Reality Exhibit \\ for Teaching Mobile Robotics Concepts in Museums}

\author{Nils Dengler \and Tim Graf \and Leif Van Holland \and Patrick Stotko \and Reinhard Klein \and Maren Bennewitz
  \thanks{All authors are with the University of Bonn, Germany. 
  M. Bennewitz, R. Klein, and N. Dengler are additionally with the Lamarr Institute for Machine Learning and Artificial Intelligence and the Center for Robotics, Bonn, Germany. 
  This work has partially been funded by the German Federal Ministry of Research, Technology and Space~(BMFTR) under the Robotics Institute Germany (RIG).}
  }%

\begin{document}
\maketitle
\thispagestyle{empty} 
\pagestyle{empty}

\begin{abstract}
Understanding autonomous mobile robots is challenging for people without robotics expertise, because key processes such as sensing, mapping, and planning are invisible.
We present RHINO-AR, an interactive Augmented Reality~(AR) museum exhibit that reintroduces the historical mobile robot RHINO into its original exhibition environment at the Deutsches Museum Bonn.
To make the autonomous behavior that once made RHINO historically significant accessible again, our previous work, \vr, reconstructed the robot and its environment in a fully virtual museum scene.
While engaging, this fully virtual presentation separated visitors from the real exhibition space and from the physical robot on display. \ours closes this reality gap by placing a virtual reconstruction of RHINO directly into the real museum space.
Implemented on a Magic Leap~2 headset using Unity, our system combines real-time environment meshing with interactive visualizations of LiDAR sensing, traversability, and path planning to make otherwise invisible robotics processes understandable to non-expert visitors. 
We evaluated \ours in a two-day museum study with 22 participants, assessing usability, technical performance, satisfaction, conceptual understanding, and preference compared to \vr. 
The results indicate that \ours was well received, conveyed the targeted navigation concepts to a non-expert audience, and was generally preferred over the VR exhibit due to its stronger physical grounding and increased realism. 
\end{abstract}

\section{Introduction}
\label{sec:intro}
Robotics has become a fundamental component of modern life, yet the underlying technologies such as localization, path planning, and obstacle avoidance remain difficult to understand for non-experts.
As a result, autonomous robots are often perceived as unpredictable ``black boxes'', potentially making public acceptance more difficult~\cite{liang2017fear, babel2022findings}. 
Robotic exhibits in museums appear suitable to address this challenge, as they allow visitors of different ages and backgrounds to explore technological systems voluntarily through observation and interaction~\cite{young2022learn, zhang2026adult}.

One of the first examples of public-facing robotics is RHINO~\cite{burgard1999experiences}, an autonomous tour-guide robot deployed in the Deutsches Museum Bonn, Germany, in 1997. 
Over six days, RHINO guided more than 2,000 visitors and demonstrated robust autonomous navigation, obstacle avoidance, and interactive tour guidance in a real museum environment.
Its deployment was one of the earliest examples of long-term autonomous robot operation in a public space and inspired later systems such as \mbox{MINERVA~\cite{thrun1999minerva}}.

\begin{figure}[t]
  \centering
  \includegraphics[width=.9\columnwidth, trim=75 15 75 15, clip]{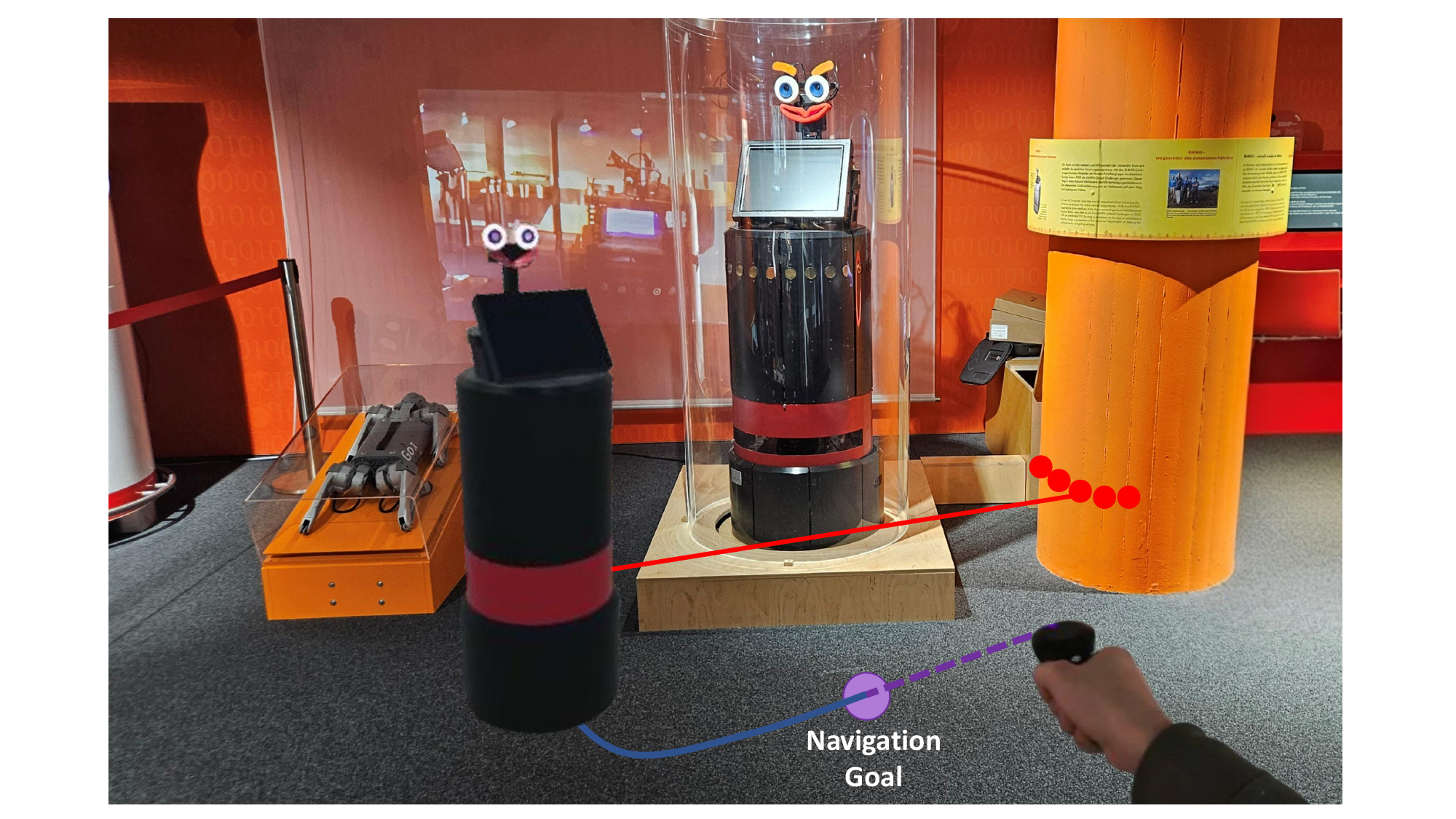}
  \caption{\ours places a virtual reconstruction of the physical RHINO (behind glass) into the real museum environment and visualizes core robot navigation concepts, including the planned path~(blue), LiDAR sensing (red), and a user-defined goal (purple).}
  \label{fig:cover}
  \vspace{-12px}
\end{figure}

Today, RHINO itself is part of the museum exhibition, but while the physical exhibit preserves the historical robot, it cannot convey the autonomous behavior that made RHINO important in the first place. 
To address this, we previously presented \vr~\cite{schlachhoff24roman}, which reconstructs both the robot and its museum environment in Virtual Reality~(VR) and allows visitors to experience RHINO's behavior in an immersive way. 
However, this approach introduced an inherent reality gap, i.e., although visitors can observe and interact with a virtual RHINO, all interaction takes place in a digital replica, detached from the actual museum space and from the physical robot on display.
While the VR exhibit was generally well received, participants identified this lack of physical grounding as a key limitation. A natural way to address this is to preserve the explanatory benefits of the virtual visualization while anchoring the experience in the real museum environment, so that visitors can relate the robot's behavior to the original exhibit and its surroundings.

Therefore, in this work, we present RHINO-AR, an Augmented Reality~(AR) museum exhibit that places a virtual reconstruction of RHINO directly into the real exhibition space (see \figref{fig:cover}). 
In contrast to VR, AR preserves the physical context of the museum while still making invisible robot processes visible, so that visitors can walk freely around the robot, place target locations, and observe how sensing, navigation, and obstacle avoidance are used in the actual museum environment.
We expect this combination of physical grounding and interactive visualization to create a more intuitive and educationally effective experience than a purely virtual presentation, and conducted a user study in the museum to investigate this assumption. 

The educational contribution of \ours is to turn otherwise invisible stages of the navigation pipeline into an observable exhibit.
We therefore formulate three learning outcomes: after using the exhibit, visitors should be able to~(i)~describe that the robot senses its surroundings through range measurements limited by line of sight and occlusion, (ii)~explain that it derives a model of traversable and blocked space from these measurements, and (iii)~recognize that motion results from planning a path toward a goal rather than from remote control.
Each visualization mode targets one of these outcomes, and our understanding questions are formulated to test them.

In summary, this paper contributes \ours, a museum exhibit that brings the historical RHINO robot back into its original museum environment and visualizes key navigation concepts such as LiDAR sensing, traversability, and path planning. In addition, we present a two-day museum study that evaluates usability, technical performance, satisfaction, conceptual understanding, and user preference.

\section{Related Work}
The increasing availability of AR- and VR-capable devices has led to a broad range of applications in entertainment, industry, medicine, and education~\cite{shatokhin2025extended}.
In education, these Extended Reality~(XR) technologies are particularly relevant for museum and classroom settings, where they make otherwise inaccessible or abstract processes directly observable~\cite{huang2025extended}.
Examples include AR-based systems for visualizing the flow of electrons in physical circuits~\cite{beheshti2017looking} and XR applications for exploring archaeological artifacts~\cite{zaia2022egyptian}.
In museums, AR has therefore been used specifically to make invisible processes visible, especially when the underlying phenomena are difficult to observe directly~\cite{yoon2014making, yoon2017augmented}.
Prior work reports that XR can increase engagement, support experiential learning, and improve retention in educational and museum contexts~\cite{lee2020experiencing, zhou2022meta}, since it complements conventional static exhibits with interactive and embodied forms of learning.
Its effectiveness, however, depends strongly on careful experience design, cognitive load management, and integration into the surrounding exhibition~\cite{chang2022ten, xu2024digitally}.

For robotics education, AR is relevant because many key processes involved in autonomous behavior are inherently hidden from human observers: concepts such as laser-based sensing, map-based localization, and path planning cannot easily be inferred from watching a static robot in a museum.

Museums have long served as spaces for informal learning outside formal education~\cite{dougan2010primary} and are therefore a suitable setting for teaching robotics to a broad audience.
Robots have been deployed there as guide robots~\cite{burgard1999experiences, thrun1999minerva}, telepresence systems, and interactive installations~\cite{maniscalco2024towards}, which can make exhibits more attractive and engaging, but whose effectiveness depends strongly on usability, embodiment, and contextual integration~\cite{germak2015robots, maniscalco2024towards}.
In parallel, robotics and AI are increasingly discussed as part of the digital transformation of museums~\cite{de2024robotics}.

XR has also been explored in robotics-specific contexts, including navigation demonstrations, safe human-robot interaction experiments, and teleoperation in remote or reconstructed environments~\cite{zacharaki2020safety, malik2021digital, youssef2023telepresence}.
In the museum context, our own \vr~\cite{schlachhoff24roman} exhibit lets visitors experience a virtual reconstruction of RHINO and its environment while learning about sensing, localization, and navigation, showing that XR can make central mobile robotics concepts accessible to a general audience and increase engagement with a historical exhibit.

\ours builds directly on \vr, but shifts the focus from virtual reconstruction toward physical grounding, i.e., it places the virtual robot into the real exhibition space, so that users observe robot behavior in the actual museum while the scene is augmented with sensor, traversability, and navigation visualizations.

\begin{figure*}[t]
  \centering
  \includegraphics[width=\textwidth, trim=0 0 0 0, clip]{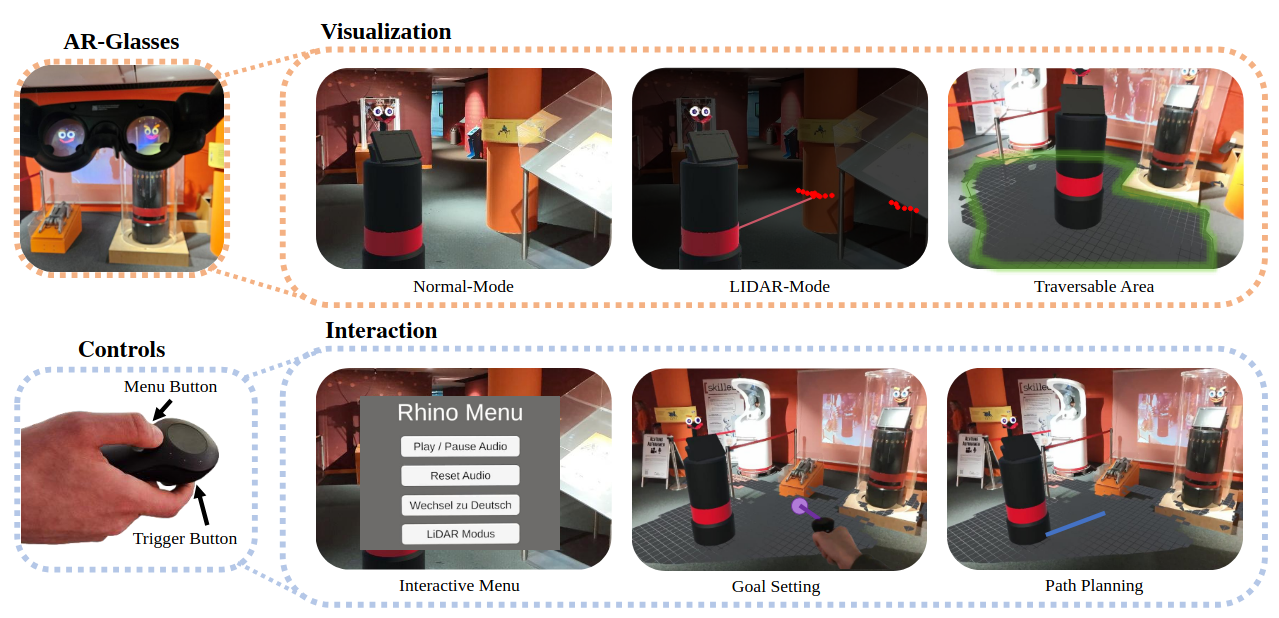}
  \caption{\ours places RHINO in the real museum environment and combines AR visualizations of standard view, LiDAR sensing, and traversability (top row) with simple controller-based interaction for menu access, goal placement, and path following (bottom row).}
\label{fig:overview}

  \vspace{-10px}
\end{figure*}

\section{Our Approach}
\subsection{System Overview}
\ours is realized in Unity~6 and built around the Niantic Lightship Augmented Reality Developer Kit~(ARDK), running on a Magic Leap~2 headset.
The design goal was to retain the educational strengths of our earlier \vr~\cite{schlachhoff24roman} exhibit while re-establishing a direct spatial connection between the visitor, the historical robot, and the real museum environment.
In contrast to the VR version, our AR system therefore does not reconstruct the exhibition digitally in advance, but continuously senses and augments the physical space as seen by the visitor, so that real movement directly translates into the experience: users can walk around the robot, follow it through the exhibition area, and inspect its behavior from multiple angles.
Since the environmental model is updated continuously, overlays and path visualizations are also occluded by dynamic obstacles such as standing or slowly moving people, which increases the perceived realism. An overview of the exhibit is shown in~Fig.~\ref{fig:overview}.

\subsection{Interaction Design}
As the target audience consists of general museum visitors with widely varying technical backgrounds, the interaction design was kept as simple as possible.
Early informal feedback showed that especially visitors with little or no prior AR or VR experience preferred a minimal control concept over richer but more complex interaction, so the system uses only two controller buttons in addition to the natural 3D pointing motion of the handheld controller. 
The menu button opens and closes the menu overlay to reset the robot, start the audio commentary, or switch between visualization modes. 
The trigger button activates menu entries and designates navigation goals, which are set by pointing at a ground region that is currently marked as navigable, with a purple pointer beam extending from the controller as a clear visual cue.
This design minimizes the need for verbal instruction and makes the interaction understandable after a very brief introduction by museum staff, as shown in the bottom half of Fig.~\ref{fig:overview}.

\subsection{Real-Time Mapping and Navigation}
\label{sec:mapping}
To place RHINO into the real museum, the application requires a continuously updated collision model of the surrounding space, which leads to two central design decisions.

First, for environment reconstruction, we compared the Magic Leap~2's native meshing system and our earlier meshing integration by Van Holland~\etalcite{van2025towards}, both relying on time-of-flight depth sensing, against the Niantic Lightship ARDK.
The depth-based variants produce more accurate geometry, but update too slowly to keep pace with a freely walking user and with passing visitors, and they are more restrictive with respect to lighting conditions.
We therefore trade geometric accuracy for update rate and build on the Lightship ARDK, since in a crowded exhibition rapid adaptation to a changing scene matters more than maximum mesh fidelity.

Second, for navigation, we convert the reconstructed surface into Unity's NavMesh representation in the ARDK-supported variant, on which a navigation agent plans the robot's motion with the A$^*$ algorithm~\cite{hart1968formal}.
This abstraction trades physical fidelity for robustness, as compared to a fully physics-based simulation, such as the PyBullet-based one used in \vr~\cite{schlachhoff24roman}, it is computationally cheaper and tolerates the noisy, frequently corrected meshes obtained in a live museum.

\subsection{Visualization of Robot Perception and Behavior}
\label{sec:viz}
A central design goal of \ours is to make invisible robot processes visible in a way that remains understandable to non-expert visitors. 
The system therefore visualizes core components of autonomous navigation directly in the real museum environment, with each mode targeting one of the three learning outcomes stated in \secref{sec:intro}, as shown at the top~of~Fig.~\ref{fig:overview}.

First, the robot's intended motion is communicated through an A$^*$~\cite{hart1968formal} path visualization drawn on the ground. Once the visitor places a navigation target, the computed path is highlighted, showing how the robot chooses a route through the available space~(\textbf{outcome~(iii)}).
Second, we visualize RHINO's LiDAR sensing, where a rotating sensor emits laser beams and estimates distances from their reflections.
Instead of displaying all simulated beams and endpoints at all times, as done in \vr~\cite{schlachhoff24roman}, we render only the detected endpoints together with a single rotating beam, which reduces visual clutter and makes the scanning principle easier to grasp in the real museum setting~(\textbf{outcome~(i)}).
Finally, an overlay of the traversable ground, computed from the dynamically reconstructed scene mesh (see \secref{sec:mapping}) as the ground regions that remain free of intersections with detected obstacles, helps visitors understand why certain target locations are accepted and others are not~(\textbf{outcome~(ii)}).

To convey both the capabilities and the limitations of range-based sensing, a second visualization mode, called \mbox{\emph{LiDAR mode}}, uses the Magic Leap~2's global dimmer to darken, but not completely remove, the user's view of the environment and retains only the sensed geometry and the detected hit points, \ie the locations at which the simulated beams intersect with surrounding surfaces. 
This mode emphasizes line-of-sight sensing, occlusion, and the dependence of robot perception on scene geometry, and proved particularly useful for explaining why robots cannot simply ``see everything'' and why their understanding of the world is only partial and sensor-dependent.

In addition to the visual overlays, a short audio explanation covers RHINO's history and navigation principles.
It is based on the content used in \vr~\cite{schlachhoff24roman}, but shortened after early feedback indicated that museum visitors preferred a concise spoken summary over a longer, multi-part explanation.
Menu and narration can be switched between German and English, and further languages can be added with little effort.
Combining visual overlays with a concise spoken explanation supports visitors who prefer either visual or verbal learning cues.


\section{Experiments}
To assess the usability, educational value, and overall reception of \ours in its intended setting, we deployed the exhibit for two days in the exhibition space of the Deutsches Museum Bonn, Germany, and conducted a user study. 
The questionnaire design is based on~\cite{schlachhoff24roman} and~\cite{leavy2014oxford}, with adaptations made to reflect the specific requirements of our use case, as described in the following.

\subsection{Study Design}

The study is designed to assess the system under realistic museum conditions, including changing lighting, moving visitors, and dynamically varying scene geometry. 

\subsubsection{\textbf{Questionnaire Overview}}
\begin{figure}[t]
  \centering
  \includegraphics[width=\columnwidth, trim=0 5 0 0, clip]{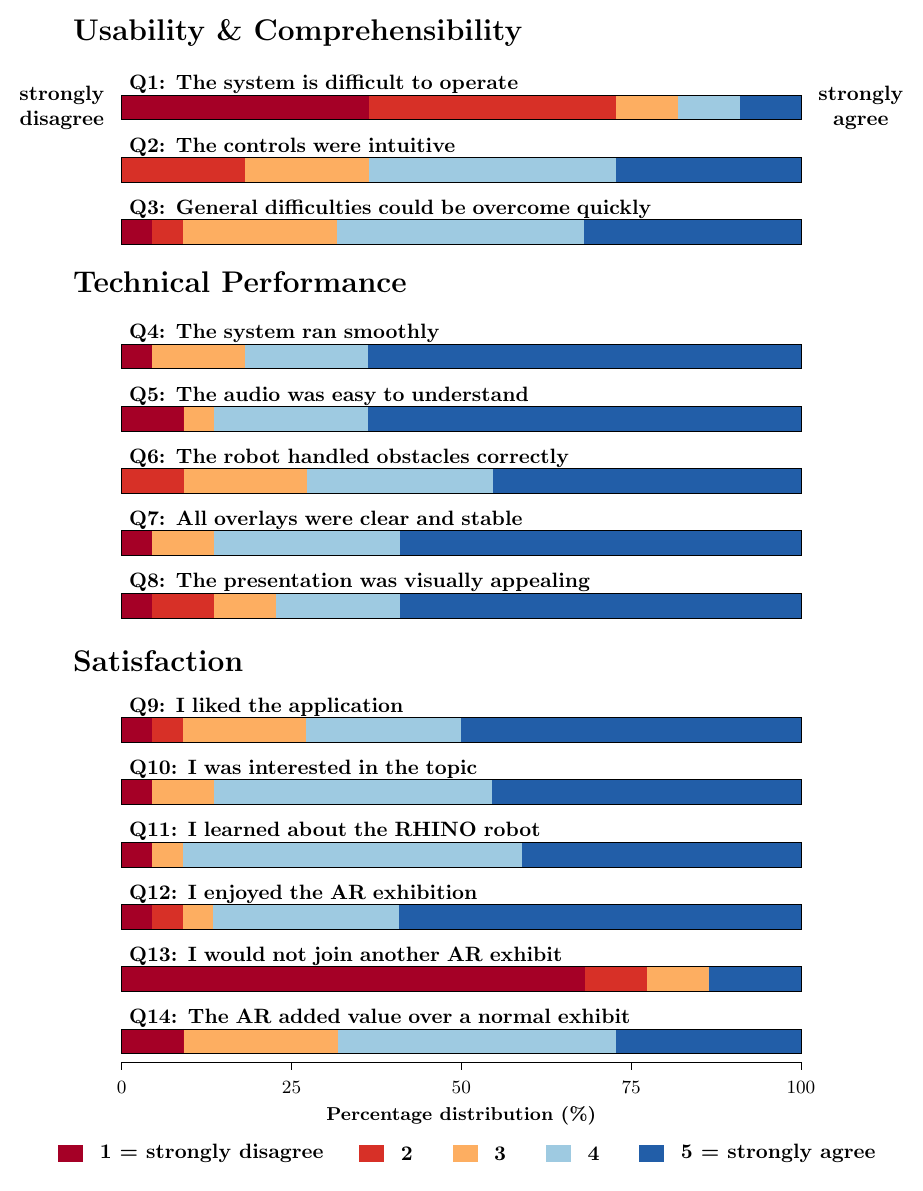}
  \vspace{-20px}
\caption{Results of the AR-specific evaluation of \ours. The bars show the distribution of participant responses for questions Q1--Q14 on a 5-point Likert scale, covering usability and comprehensibility, technical performance, and overall satisfaction. 
Overall, the distributions indicate a predominantly positive assessment of the system, with especially favorable ratings for technical performance and user satisfaction.
Note that Q1 and Q13 are negatively phrased items and must therefore be interpreted inversely.}
  \label{fig:ar_results}
  \vspace{-12px}
\end{figure}

The questionnaire followed the same structure as in~\cite{schlachhoff24roman} and allows both an \mbox{AR-specific} evaluation and a direct comparison to the earlier VR exhibit.
Participants first reported demographic information and prior familiarity with AR/VR technologies and robotics.
The AR exhibit itself was then assessed using three groups of questions, as shown in \figref{fig:ar_results}: \textbf{usability and comprehensibility}~\mbox{(three questions)}, \textbf{technical performance}~\mbox{(five questions)}, and \textbf{satisfaction}~\mbox{(six questions)}. 

In contrast to~\cite{schlachhoff24roman}, participants additionally answered a small set of understanding questions that target the learning outcomes defined in \secref{sec:intro}. 
Furthermore, to explicitly compare both systems, all 11 participants of the second day were presented with both systems and completed a direct comparison between \ours and \vr~\mbox{(nine questions, \figref{fig:comparison})}.
Finally, one open-form question asked for general feedback.

All closed-form AR evaluation items used a 5-point Likert scale ranging from ``strongly disagree''~($1$) to ``strongly agree''~($5$), where Q1 and Q13 are phrased negatively on purpose to identify inattentive or inconsistent responses.
In the direct comparison section, participants selected whether AR, VR, or both systems matched the respective statement, and the understanding questions used binary checkboxes.

\subsubsection{\textbf{Procedure}}
In total, 22 participants took part in the study, with~11~participants on each day, representing a heterogeneous museum audience with varying levels of prior familiarity with robotics and XR technology.
Participants were approached when showing interest in the exhibit, were given a short verbal introduction, and completed the demographic and prior knowledge section.
They were then fitted with the Magic Leap~2 headset and, after a brief explanation of the two controller inputs, explored \ours freely: they could inspect the virtual RHINO robot from different viewpoints, place target locations, observe the displayed path planning behavior, and switch to the dedicated LiDAR mode.
Afterwards, they completed the remaining questionnaire sections. On the second day, participants additionally experienced \vr and completed the direct comparison and overall preference sections.
A \mbox{Mann-Whitney~$U$~test} on the collected response variables found no significant differences between the two study days ($p > 0.05$), indicating that the responses originate from the same distribution.

\subsubsection{\textbf{Demographics}}
The demographic questionnaire collected age and gender anonymously.
Participation was open to all interested visitors who were of legal age or had permission from a legal guardian, which reflects the natural audience of a public museum exhibit and resulted in a heterogeneous sample.
On the first study day (Saturday), participants had a mean age of 28.9~years ($\pm 18.9$), with 4~female and 7~male participants, while on the second day (Thursday), the mean age was 45.3~years ($\pm 17.2$), with 5~female and 6~male participants.
This difference can plausibly be explained by the visitor composition on weekends and weekdays. On Saturday, participation often involved parents together with their children, resulting in a younger and more variable sample, whereas the weekday sample consisted predominantly of adult visitors. 
Overall, the sample reflects the diversity of the intended target group, namely museum visitors with varying ages and levels of prior technical familiarity.
With 11~participants per day, however, it is too small to support statistically meaningful analyses of acceptance by age group or gender, so we report the demographics only to characterize the sample rather than as an independent variable.

\subsubsection{\textbf{Prior Experience}}
On three 5-point Likert items, participants rated whether they had already used VR or AR glasses, whether they had already used other VR or AR museum applications, and whether they were familiar with robots such as robotic vacuum cleaners, yielding mean ratings of \mbox{$2.95 \pm 1.70$}, \mbox{$3.05 \pm 1.76$}, and \mbox{$2.59 \pm 1.53$}, respectively.
Averaged across all items, prior experience was moderate ($M=2.86$, $SD=1.43$) but varied substantially between individuals, so that the sample included both technically experienced visitors and participants with little prior familiarity, as intended for a public museum exhibit.
\vspace{-5px}
\subsection{Results}
\figref{fig:ar_results} summarizes the AR-specific evaluation and illustrates the distribution of responses for all questionnaire items.
A two-sided one-sample t-test showed that all items differed significantly from the neutral midpoint of the Likert scale ($p<0.05$), indicating an overall positive assessment across usability, technical performance, and satisfaction.


\subsubsection{\textbf{Usability and Comprehensibility}}
The first AR-specific block assessed ease of use and comprehensibility using three questionnaire items.
The negatively phrased item Q1 (\textit{``The system is difficult to operate''}) received an inverted score of $3.82 \pm 1.30$ (originally $2.18 \pm 1.30$), indicating that participants generally did not perceive the system as difficult to operate, while Q2 and Q3 were rated positively with $3.73 \pm 1.08$ and $3.86 \pm 1.08$, respectively.
These results suggest that the interaction scheme was understandable and usable even for visitors with little or no previous AR experience. 

\subsubsection{\textbf{Technical Performance}}
Questions Q4--Q8 assessed the perceived technical performance. 
The mean ratings were consistently high, ranging from $4.09 \pm 1.02$ for obstacle handling to $4.36 \pm 1.05$ for smooth overall execution, with $4.18 \pm 1.22$ for visual appeal. 
These values indicate that participants experienced the system as technically convincing despite the challenging museum setting, and the high ratings for smoothness and overlay stability suggest that the dynamic meshing and visualization pipeline were sufficiently robust for public-facing deployment.

\subsubsection{\textbf{Satisfaction}}
Visitor satisfaction was measured using questions Q9--Q14.
All positively phrased items were rated favorably, with mean scores from $3.77 \pm 1.15$ to $4.32 \pm 1.09$, and the negatively phrased item Q13 (\textit{``I would not join another AR exhibit''}) received a high inverted score of $4.18 \pm 1.44$~(originally $1.82 \pm 1.44$), indicating a high willingness to engage with similar exhibits in the future.
Taken together, participants found the application enjoyable, educational, and valuable in the museum context. 


\subsubsection{\textbf{Understanding of RHINO's Navigation}}
To assess whether the exhibit conveyed the intended learning outcomes, the questionnaire
included four dedicated understanding questions. 
In particular, these binary choice questions focused on RHINO's sensing and path planning as well as on what RHINO was originally used for. 
Across all 22 participants, the mean correctness score was $0.83$ on a scale from $0$ to $1$: 12~participants achieved a fully correct score of $1.0$, the remaining ones obtained partial scores of $0.75$ $(n=7)$, $0.50$ $(n=1)$, and $0.25$ $(n=2)$, and no participant received a score of $0$. 
These results suggest that most participants reached the targeted learning outcomes after a single short interaction. 
Given the small sample and the absence of a pre-test baseline, however, they should be read as an encouraging indication of immediate comprehension rather than as a measurement of learning gain, as discussed in \secref{sec:limitations}.

\begin{figure}[t]
  \centering
  \includegraphics[width=\columnwidth, trim=0 5 0 0, clip]{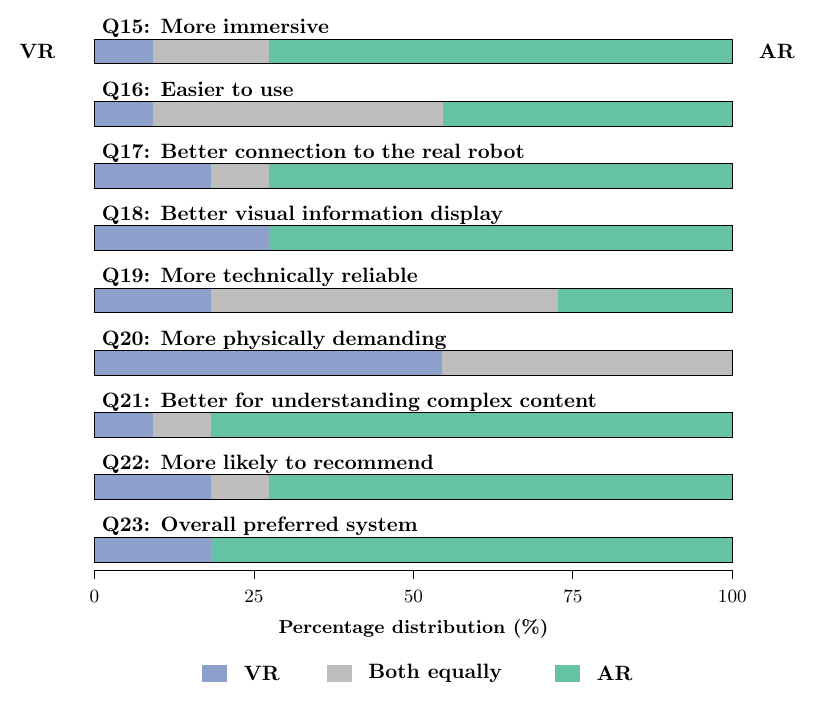}
  \vspace{-20px}
  \caption{Results of the direct comparison between \ours and \mbox{\vr~\cite{schlachhoff24roman}}. The bars show the distribution of participant responses to questions Q15--Q23, where participants selected which system better matched the respective statement. Overall, the distributions indicate a clear preference for \ours in most categories, particularly with respect to immersion and support for understanding complex content.}
  \label{fig:comparison}
  \vspace{-15px}
\end{figure}

\subsubsection{\textbf{Comparison to \vr}}
On the second study day, 11 participants directly compared \ours and \mbox{\vr}.
The corresponding results are shown in \figref{fig:comparison}. 
In the direct comparison section (Q15--Q20), AR is preferred by a clear majority for immersion, connection to the real robot, and visual information display.
In contrast, VR was perceived as physically more demanding, which may be explained by the stronger sensory isolation of the headset-based virtual environment and the increased effort required to orient oneself and move confidently in a completely virtual~space.

The overall preference section (Q21--Q23) showed an even clearer tendency toward AR, as most participants stated that they preferred the AR system overall, would be more inclined to recommend it, and found that it better helped them understand complex content.
This indicates that the physical grounding of AR in the real museum environment was highly appreciated, while both media retain different strengths: AR benefits from co-location with the historical exhibit, whereas, as stated in~\cite{schlachhoff24roman}, VR remains valuable when a fully controlled experience is desired.

\section{System Trade-Offs, Limitations,\\ and Future Work}
\label{sec:limitations}
\ours deliberately prioritizes robustness and clarity over strict physical realism. Since headset-based meshing only captures successfully detected surfaces, transparent materials such as glass railings or windows may be reconstructed incompletely. A fully physics-based robot simulation was moreover avoided because noisy or delayed geometry updates can leave the robot stuck, knocked over, or hard to recover; the final navigation abstraction instead keeps the interaction stable in a public museum setting.

A second set of limitations concerns the scope of our evaluation. With 22~participants for the AR-specific items and only 11~for the direct comparison to \vr, the study gives a first indication of how the exhibit is received, but is too restrictive to analyze acceptance across age groups or genders and does not support strong claims about educational effectiveness. The understanding questions were moreover answered directly after a single short walk-up interaction and without a pre-test baseline, so they capture immediate comprehension rather than learning gain or long-term retention. Finally, recruiting among visitors who already showed interest in the exhibit introduces a self-selection bias toward a technology-affine audience.

Future work will therefore extend the deployment over a longer period to obtain a larger and more diverse sample, in particular for the comparison with \vr, and add a pre-/post-test design with a delayed follow-up to measure the actual learning gain. Beyond this, we plan to investigate gamification elements, such as small navigation tasks or obstacle-avoidance challenges, to further increase engagement and reinforce the conveyed concepts.

\section{Conclusion}
In this paper, we presented \ours, an augmented reality museum exhibit that reintroduces the historical mobile robot RHINO into its original exhibition environment at the Deutsches Museum Bonn. 
By anchoring a virtual reconstruction of the robot in the real museum space, \mbox{\ours} preserves the physical context of the exhibit while making otherwise invisible robotics processes such as sensing, traversability estimation, and path planning visible in an intuitive way, thereby addressing the reality gap of our \vr~\cite{schlachhoff24roman} exhibit.
In our user study with 22 participants, the system was rated positively with respect to usability, technical performance, and overall satisfaction, and the understanding questions suggest that most visitors reached the intended learning outcomes, although a larger and longitudinal study is needed to quantify the actual learning gain.
In the direct comparison, participants generally preferred the AR version, especially due to its stronger physical grounding and increased perceived realism in relation to the historical exhibit.

%


\bibliographystyle{IEEEtran}

\bibliography{bibliography}

\end{document}